\documentclass[runningheads]{llncs}
\usepackage{graphicx}

\usepackage{tikz}
\usepackage{comment}
\usepackage{amsmath,amssymb} 
\usepackage{booktabs}
\usepackage{multirow}
\usepackage{hyperref}
\usepackage{color}

\usepackage{graphicx}
\usepackage{float}
\usepackage{subfig}
\usepackage[accsupp]{axessibility}  

\newcommand{\ie}{\textit{i}.\textit{e}.}
\newcommand{\eg}{\textit{e}.\textit{g}.}

\makeatletter
 \def\@fnsymbol#1{\ensuremath{%
    \ifcase#1
    \or 
      \dagger
    \or 
      \ddagger
    \or 
      \mathsection
    \or 
      \mathparagraph
    \else 
      \@ctrerr  
    \fi}}   
\makeatother   

\begin{document}
\pagestyle{headings}
\mainmatter
\def\ECCVSubNumber{2948}  

\title{Spatial and Visual Perspective-Taking via View Rotation and Relation Reasoning for Embodied Reference Understanding} 


\titlerunning{REP for Embodied Reference Understanding}

\author{Cheng Shi\inst{1} \and
Sibei Yang\inst{1,2}\thanks{Corresponding author}}


\institute{School of Information Science and Technology, ShanghaiTech University
\\
\and
Shanghai Engineering Research Center of Intelligent Vision and Imaging\\
\email{\{shicheng, yangsb\}@shanghaitech.edu.cn
}}
\maketitle

\begin{abstract}
Embodied Reference Understanding studies the reference understanding in an embodied fashion, where a receiver requires to locate a target object referred to by both language and gesture of the sender in a shared physical environment. Its main challenge lies in how to make the receiver with the egocentric view access spatial and visual information relative to the sender to judge how objects are oriented around and seen from the sender, \ie, spatial and visual perspective-taking. In this paper, we propose a \textbf{REasoning from your Perspective} (REP) method to tackle the challenge by modeling relations between the receiver and the sender as well as the sender and the objects via the proposed novel view rotation and relation reasoning. Specifically, view rotation first rotates the receiver to the position of the sender by constructing an embodied 3D coordinate system with the position of the sender as the origin. Then, it changes the orientation of the receiver to the orientation of the sender by encoding the body orientation and gesture of the sender. Relation reasoning models both the nonverbal and verbal relations between the sender and the objects by multi-modal cooperative reasoning in gesture, language, visual content, and spatial position. Experiment results demonstrate the effectiveness of REP, which consistently surpasses all existing state-of-the-art algorithms by a large margin, \ie, +$5.22\%$ absolute accuracy in terms of Prec@$0.5$ on YouRefIt. Code is available\footnote[1]{\url{https://github.com/ChengShiest/REP-ERU}}.
\end{abstract}

\section{Introduction}
\label{sec:intro}

\begin{figure}[htb]
\centering
\includegraphics[width=0.8\columnwidth]{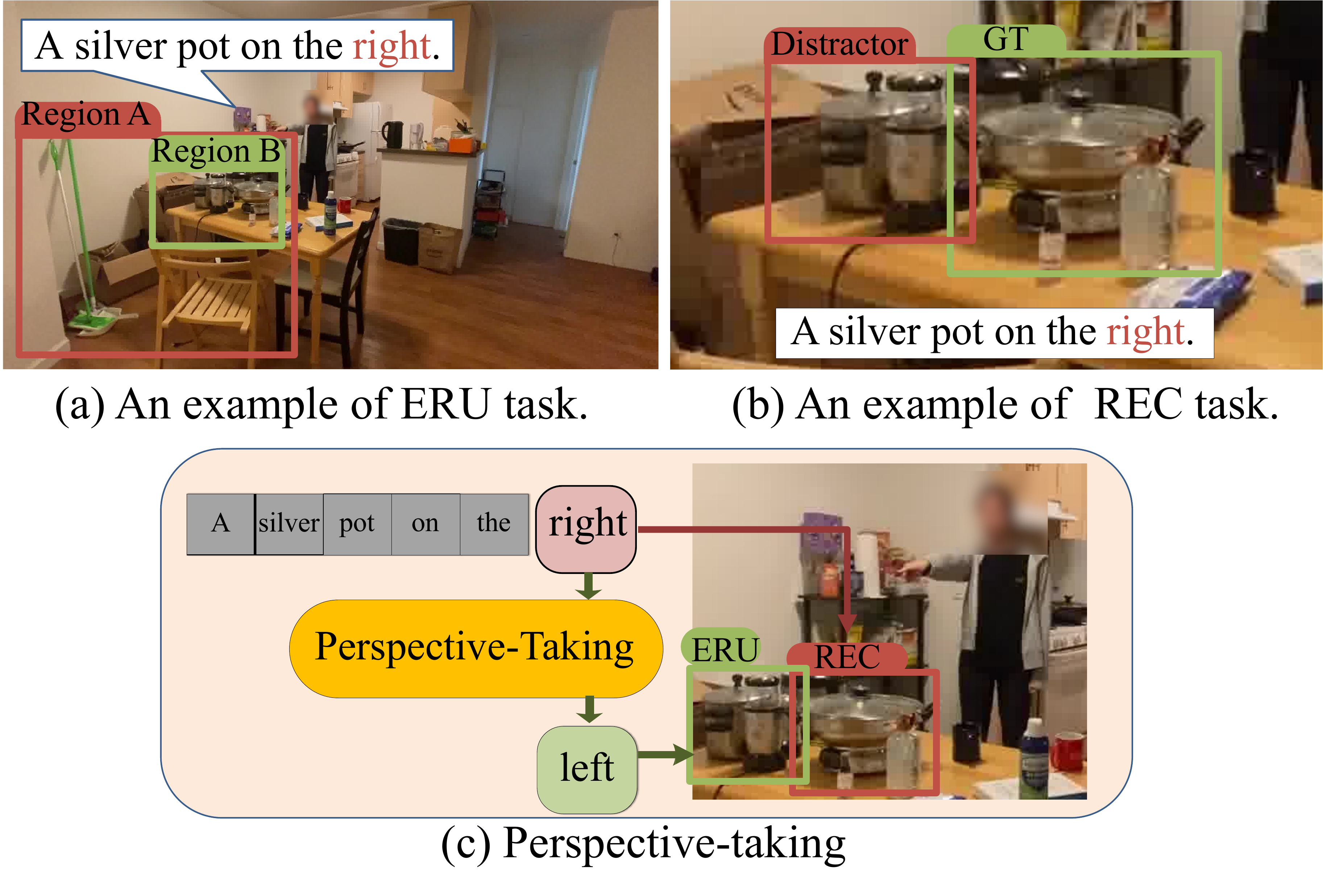}
\caption{The difference between Embodied Reference Understanding (ERU) and traditional Reference Expression Comprehension (REC). Figure (a) and (b) are two examples of ERU and REC, respectively. In (a), the person (\ie, sender) in the image gives the description, while an annotator generates the description according to the image in (b). Figure (c) shows that the two tasks differ in the localization of the target objects according to the same language description due to the perspective-taking challenge.}
\label{fig1}
\end{figure}

Reference understanding, recognizing referents (\eg, target objects) which are referred to by interlocutors in a shared environment, helps to establish common ground in human communication~\cite{chen2021yourefit}. Referring Expression Comprehension (REC)~\cite{mao2016generation,yu2016modeling,hu2017modeling,yu2018mattnet,yang2019cross}, a reference understanding task in computer vision community, aims at learning a receiver to detect the referent from an image corresponding to a natural language sentence generated by a sender. In REC, the receiver and the sender recognize the referent of the image from the same viewpoint, \ie, the camera viewpoint. An example of REC is shown in Fig.\ref{fig1}(b). 
Instead of considering cooperative communication~\cite{stacy2020intuitive} with human-in-the-scene, REC emphasizes the joint understanding of visual and language cues. 

To facilitate reference understanding in an embodied environment, Embodied Reference Understanding (ERU)~\cite{chen2021yourefit} with benchmark and dataset (\ie, YouRefIt) is proposed recently. ERU task mimics the referring process of human communication in an embodied manner, in which the sender and the receiver are in the same physical space but they observe the referent from different viewpoints. An example of ERU is shown in Fig.\ref{fig1}(a). The sender describes the ``A silver pot on the right" from her perspective, and the receiver requires to locate the pot in the receiver's first-person image. Both sender's language description and gesture to the referent are included in YouRefIt~\cite{chen2021yourefit} because people often jointly use these verbal and nonverbal forms to refer to an object.

Spatial and visual perspective-taking~\cite{surtees2013similarities,clinton2018gaining} is a challenging but essential factor to address ERU, where it requires the receiver to access spatial and visual information relative to the sender to judge how objects are oriented around and seen from the sender.
We claim that the sender's position, gesture information, and corresponding visual cues in the 3D scene are vital to achieving spatial and visual perspective-taking. Specifically, the sender's 3D position in the physical environment and body orientation implied in visual appearance can indicate the rough area that the sender pays attention to. For example, as shown in Fig.\ref{fig1}(a), the ``region A" facing the sender is more likely to be paid attention to by the sender. 
By cooperating with the gesture, we can estimate a more accurate, nonverbal-aware attention distribution of regions, \ie, the ``region B" around the table will be attended more. However, existing reference understanding methods~\cite{yang2019fast,yang2020improving,chen2021yourefit} either neglect the perspective-taking challenge or address it by simply fusing a gesture map with visual cues, which cannot achieve satisfactory performance.
How to cooperate language cues with the perspective-taking to implement the reference understanding with verbal input serves as a crucial problem. 
The language descriptions in ERU usually contain two types of information, \ie, the appearance of the referent and spatial relation between the referent and the sender. Therefore, we must combine both types of information with perspective-taking to perform relation reasoning from the sender's viewpoint, which is also different from the REC task with viewpoint-only compositional reasoning~\cite{yang2020graph,yang2020improving}. 
The appearance information with open-vocabulary category and attribute description helps locate the candidate objects from the attention regions. For example, two pots shown in Fig.\ref{fig1}(c) are figured out through the category description ``pot" and attribute description ``silver". Moreover, the spatial relation cues from the language description such as ``front" and ``right" represent the relative spatial relationship between the referent and the sender. 
For example, as shown in Fig.\ref{fig1}(c), the pot with a green bounding box will be identified because it is on the ``right" of the sender from the sender's perspective. 

In this paper, we propose a one-stage \textbf{REasoning from your Perspective} (REP) network to address the perspective-taking and multimodal cooperation challenges in ERU. REP explictly performs the relation modeling between the receiver and the sender as well as the sender and the objects via the proposed 3D view rotation and relation reasoning modules. Specifically,  (1) \textit{REP captures the relation between the receiver and the sender via the 3D view rotation module in the following two steps.} First, it rotates the receiver into the sender's position by estimating the depth from the image and constructing an embodied 3D coordinate system with the sender's positon as the origin. 
Second, it encodes the body orientation and gesture of the sender in the 3D coordinate system to the body language vector by fusing the visual and spatial cues, including the image, gesture, and coordinate information.
The body language vector represents the orientation from the sender's viewpoint to referent in the 3D coordinate system. 
(2) \textit{Next, REP performs relation reasoning between the sender and the objects by utilizing both the verbal and nonverbal cues.} First, REP obtains the spatial attention between the body language vector and the embodied 3D spatial coordinates of all the pixels in the image. The attention distribution indicates the area where the sender faces. Second, in that area, REP performs nonverbal reasoning and verbal reasoning to find the precise region where the sender points to and describes. The nonverbal reasoning models the relations among different regions and the sender via the self-attention mechanism~\cite{vaswani2017attention}, while verbal reasoning stepwisely performs language-conditional normalization~\cite{yang2020improving,chen2021yourefit,perez2018film}. Finally, REP combines both nonverbal reasoning and verbal reasoning to predict the referent.

In summary, this paper makes four major contributions:
\begin{itemize}
    \item To the best of our knowledge, we are the first to explicitly model the relation between receiver and sender (\ie, receiver-sender relation) as well as the relation between sender and objects (\ie, sender-object relation) to address the spatial and visual perspective-taking and multimodal cooperation challenges in Embodied Reference Understanding (ERU).
    \item We propose a 3D view rotation module to rotate the receiver to sender's position and encode the direction from the sender's viewpoint to the referent for receiver-sender relation modeling, making the receiver adapts to the sender's spatial position, gesture, and body orientation. 
    \item We propose a relation reasoning module to perform verbal and nonverbal reasoning for sender-object relation modeling, which meets the requirement of ERU for multimodal cooperation. 
    \item Experimental results demonstrate that the proposed REP not only significantly outperforms existing state-of-the-art methods but also generates explainable visual evidence of stepwise reasoning.

\end{itemize}

\section{Related Work}\label{sec:relatedwork}

\subsection{From Referring Expression Comprehension to Embodied Reference Understanding} 

Referring Expression Comprehension~\cite{mao2016generation,yu2016modeling} aims at detecting the referent object from an image according to a natural language description. Works in referring expression comprehension can be roughly divided into two types, \ie, two-stage and one-stage methods.
Compared to the proposal generation and then the prediction of two-stage methods~\cite{nagaraja2016modeling,hu2017modeling,zhang2018grounding,yu2018mattnet,liu2019improving,yang2019cross,wang2019neighbourhood,yang2020relatinship-embedded,yang2019dynamic}, one-stage methods~\cite{yang2019fast,yang2020improving,chen2018real,sadhu2019zero,liao2020real,luo2020multi,yang2020propagating} directly predict the referent by regressing coordinates of it. 
FAOA~\cite{yang2019fast} fuses text features of the description into YOLOv3 detector~\cite{redmon2018yolov3} to make referring expression comprehension one-stage. To ground complex descriptions, ReSC~\cite{yang2020improving} improves FAOA by proposing a sub-query construction to refine text-conditional visual representation recursively.

Referring expression comprehension mainly focuses on jointly understanding the vision and language, 
which limits the application of reference understanding in daily embodied scenes: the sender describes the referent to another people (\ie, the receiver) in the shared physical space~\cite{fan2021learning,wu2021communicative}.
To extend referring expression comprehension to embodied scenes, Chen~\textit{et al.}~\cite{chen2021yourefit} present a new challenging reference understanding task called Embodied Reference Understanding (ERU) and collect its corresponding benchmark dataset, \ie, YouRefIt. In addition to language descriptions, gestural information is included in YouRefIt because people often use both natural language and gestures to refer to an object in the embodied setting. 
To encode the nonverbal gestural information for prediction, Chen~\textit{et al.} introduce a Part Affinity Field (PAF) heatmap~\cite{cao2019openpose} and a saliency heatmap~\cite{kroner2020contextual} and fuse them with verbal language cues and visual features. Their one-stage architecture and fusion method are based on ReSC.

Although jointly encoding multiple modalities (natural language, gestures, and images) for prediction, existing state-of-the-art methods (referring expression comprehension methods~\cite{yu2018mattnet,yang2019fast} and the embodied multimodal framework~\cite{chen2021yourefit}) fail to address a crucial challenge in ERU, \ie, visual perspective-taking~\cite{batson1997perspective,qiu2020human,chen2021yourefit}. Visual perspective-taking is the receiver's awareness and ability to imagine how the sender sees things and describe the referent from their perspective. To solve these issues, we first transfer the receiver's perspective to the sender's one via a embodied 3D coordinate construction and body orientation estimation of the sender. Then, we perform spatial and visual reasoning between objects according to the langauge and gesture cues.


\subsection{Single Image Depth Estimation}
Single image depth estimation~\cite{eigen2014depth} aims at estimating a dense depth map from a single RGB image. Occlusion between objects and perspective, including size cue and texture gradient, are keys for monocular depth estimation~\cite{mertan2021single}, and several learning models~\cite{eigen2015predicting,fu2018deep,bhat2021adabins} based on these cues are proposed. Apart from learning the depth estimation individually, some works jointly solve single image depth estimation task with other similar tasks such as semantic segmentation~\cite{ladicky2014pulling}, surface normal estimation~\cite{qi2018geonet} and contour estimation~\cite{xu2018pad}. 

Depth estimation from a single image is also introduced in vision-and-language tasks, which need depth information to reduce ambiguity in resolving scene geometry. Banerjee \textit{et al.}\cite{banerjee2021weakly} propose to utilize the depth information estimated by the off-the-shelf depth estimator AdaBins~\cite{bhat2021adabins} as the weak supervision sign to help learn the relative spatial position between objects for the visual question answering task. 
AdaBins adopts a transformer-based architecture that divides the depth range into scene-relevant bins adaptively and estimates depth values as linear combinations of these bin centers. In this paper, we also use AdaBins to extract 3D scene geometry from a single image. Different from previous methods, we cooperate the scene geometry with gesture cues and position information of the sender to estimate spatial attention distribution and spatial relationship between the sender and objects, respectively. 

\subsection{Relation Reasoning in Reference Understanding}

Relation reasoning, the ability to understand and perform reasoning of spatial and visual relations between visual regions, is explored in the related topics of reference understanding, such as referring expression comprehension~\cite{hu2017modeling,yu2018mattnet,wang2019neighbourhood,yang2020graph} and visual question answering~\cite{johnson2017clevr,hudson2018compositional,banerjee2021weakly}. These works mainly resort to neuro-symbolic methods, attention mechanisms, or graph-based methods to perform compositional relation reasoning. Specifically, neuro-symbolic methods first extract symbolic representations and then execute neuro-symbolic programs~\cite{mao2019neuro,yi2018neural} based on the representations, while graph-based methods capture the relation context via graph neural networks~\cite{kipf2016semi}. However, these methods cannot be utilized to embodied reference understanding directly. On one hand, natural language sentences on the embodied settings are much shorter than those of other reference understanding tasks, the few relation-relevant language cues should be combined with the gestures to guide the relation reasoning. On the other hand, as the sentences are described by the sender whose perspective is different from the receiver, the relation reasoning should be adaptive to the perspective-taking challenge. In this case, we convert the image coordinate to a sender-centric one and perform spatial reasoning with language and gesture cues on converted coordinates.

\section{REasoning from your Perspective}
\begin{figure*}[htb]
\centering
\includegraphics[width=\textwidth]{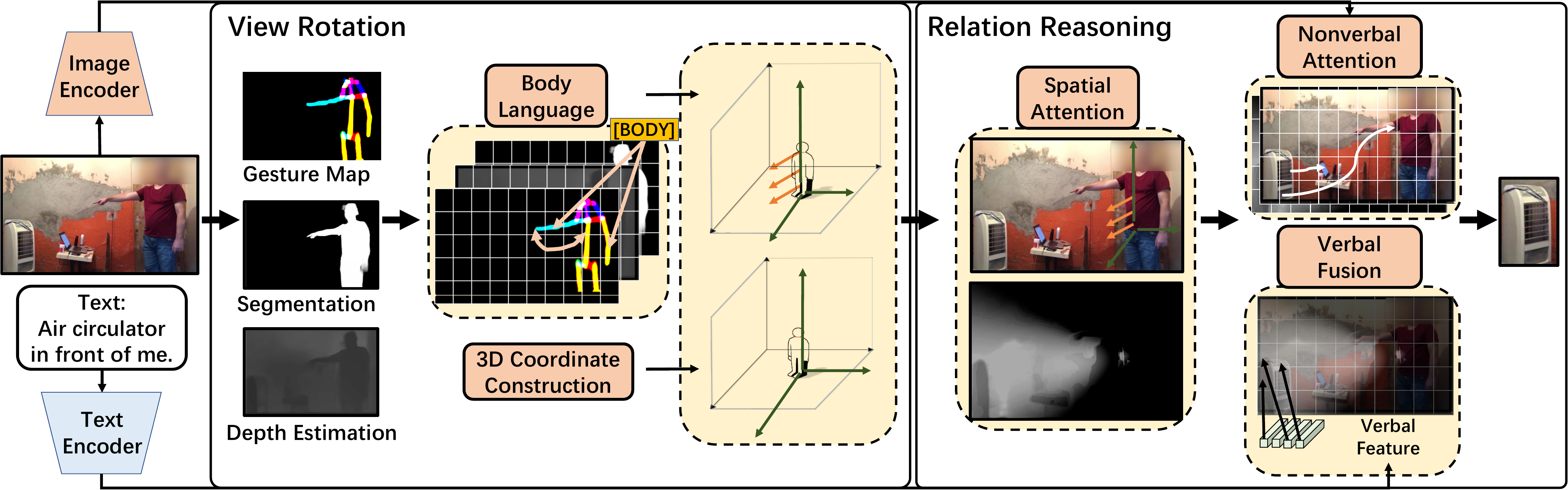}
\caption{\textcolor{black}{An overview of our Reasoning from Your Respective (REP) model. In 3D view rotation, REP first uses the depth estimation to get the 3D coordinate map and converts it to an embodied 3D coordinate map by taking the sender's position as the origin. Then, the body language vector is encoded from the visual feature map, the gesture map and the depth estimation to represent the gesture and orientation information. In Relation Reasoning, to locate the spatial area where the
sender faces, REP computes the spatial attention between the learned body language vector and spatial coordinates of all the pixels in the image. Then, in that area, REP performs the noverbal gesture attention and verbal fusion to find the precise region where the sender points to and describes. In the end, according to the precise region, REP generates a box prediction to the referent.}}
\label{fig2}
\end{figure*}

We propose a REasoning from your Perspective (REP) model to tackle Embodied Reference Understanding (ERU) task. As shown in Fig.\ref{fig2}, REP locates the referent via the 3D view rotation module and relation reasoning module. 
First, the 3D view rotation module (in section~\ref{sec:3DVR}) rotates the receiver to the sender's position by constructing an embodied 3D coordinate system and encodes the direction from the sender to the referent by learning the body language vector. Next, the relation reasoning module (in section~\ref{sec:RR}) models the relations between visual regions and the sender by cooperating with the spatial attention, nonverbal gesture reasoning and verbal fusion. Finally, we introduce the loss to train our REP in section~\ref{sec:loss}.

\subsection{3D View Rotation}\label{sec:3DVR}

We model the relation between the receiver and the sender to make the receiver could access spatial and visual information relative to the sender by constructing an embodied 3D coordinate system and learning the sender's body language representation. To construct the embodied 3D coordinate system, we first combine the raw image coordinate with the estimated depth from the image to obtain the 3D spatial information and then construct the coordinate system by setting the origin as the sender's position. Next, we estimate the direction from the sender to the referent by learning the sender's body language representation from spatial, gesture, and visual information. 

\subsubsection{Embodied 3D Coordinate System Construction}\label{sec:E3CSC}

As the referring action takes place in the 3D physical environment, the gesture and language cues relevant to the reference understanding and reasoning are based on the 3D scene. To better align the gesture and langauge cues with the spatial information, we thus construct a 3D coordinate system via depth estimation from the 2D image. Given an input image $\boldsymbol{I}$ with size of $H_I \times W_I$, we first obtain the normalized image coordinate map $\boldsymbol{P}_I \in \mathbb{R}^{H_I \times W_I \times 2}$, where $\boldsymbol{P}_I(x,y)$ is the normalized coordinate $(\frac{x}{H_I}, \frac{y}{W_I})$ of the pixel at the position $(x, y)$ in the image. Then, we estimate the image's dense depth map $\boldsymbol{P}_D \in \mathbb{R}^{H_I \times W_I}$ by using the AdaBins estimator~\cite{bhat2021adabins} trained on the indoor dataset NYU~\cite{silberman2012indoor} and then concatenate the normalized depth map $\boldsymbol{P}_D$ with the normalized image coordinate map $\boldsymbol{P}_I$ to get the 3D coordinate map $\boldsymbol{P} \in \mathbb{R}^{H_I \times W_I \times 3}$.

In order to access scene information relative to the sender, we convert the 3D coordinate map $\boldsymbol{P}$ to a sender-centric one by taking the sender's position as the origin. First, we estimate the position $\boldsymbol{p} \in \mathbb{R}^{3}$ of the sender by obtaining the person segmentation mask $\boldsymbol{A}_{sender} \in \{0,1\}^{H_I \times W_I \times 1}$ of the sender via ${\rm U^2}$-Net~\cite{qin2020u2} and setting the position $\boldsymbol{p}$ as the average coordinates of pixels that belong to the sender. 
Then, we establish the embodied 3D coordinate system by calculating the coordinates of pixels relative to the sender, and the embodied coordinate map $\boldsymbol{P}_r \in \mathbb{R}^{H_I \times W_I \times 3}$ is computed as follows,
\begin{equation}\label{eq1}
\boldsymbol{P}_r=\boldsymbol{P}-Tile(\boldsymbol{p}),
\end{equation} 
where $Tile(\cdot)$ means to tile a vector to produce a map with the size of $H_I \times W_I \times 3$.

\subsubsection{Body Language Representation}~\label{sec:BLR}
As the referent is usually in the region where the sender faces, the sender's body orientation indicates the direction from the sender to the region. To capture the body orientation, we first extract and fuse body-relevant multimodal information, including the spatial coordinate, gesture, and visual appearance, and then models the intra-relation among different parts of the sender's body. First, we extract visual feature map $\boldsymbol{S}_v \in \mathbb{R}^{H \times W \times C}$ and part affinity field map $\boldsymbol{S}_{gesture} \in \mathbb{R}^{H \times W \times 3}$~\cite{cao2019openpose,chen2021yourefit} from the image to encode the sender's visual apperance and gesture, respectively. Second, we fuse the visual features $\boldsymbol{S}_v$, gesture features $\boldsymbol{S}_{gesture}$, and their corresponding 3D spatial coordinates $\boldsymbol{P}_r$ to obtain the multimodal feature map $\boldsymbol{M} \in \mathbb{R}^{H \times W \times C}$, which is formulated as, 
\begin{equation}\label{eq2}
\begin{aligned}
\boldsymbol{M} &= Conv_{1\times 1}([\boldsymbol{S}_v; \boldsymbol{S}_{gesture}; AvgPool(\boldsymbol{P}_r)]),
\end{aligned}
\end{equation}
where $AvgPool(\cdot)$ is to downsample the feature map to the size of ${H \times W}$ via the average pooling operation, and $[;]$ and $Conv_{1\times1}(\cdot)$ refers to the concatenation operation and convolutional layer with kernel size $1\times1$, respectively. 

To force the relation modeling focus on the regions of the sender's body, we further fuse the sender's segmentation mask $\boldsymbol{A}_{sender}$ with the multimodal feature map $\boldsymbol{M}$. The fused body feature map $\boldsymbol{M}_{body} \in \mathbb{R}^{H \times W \times C}$ is computed as follows,
\begin{equation}
    \boldsymbol{M}_{body} = \boldsymbol{M} \odot AvgPool(\boldsymbol{A}_{sender}),
\end{equation}
where $\odot$ is the element-wise multiplication.

Next, we capture the intra-relation among different parts of the sender's body to predict the sender's body orientation. Specifically, we flatten the multimodal feature map $\boldsymbol{M}_{body}$ into a squence of $H \times W$ tokens $[\boldsymbol{M}_{body}^{(1,1)}, \boldsymbol{M}_{body}^{(1,2)}, ..., \boldsymbol{M}_{body}^{(H,W)}]$ and apply a stack of transformer encoder layers~\cite{vaswani2017attention} to build the global correlation among the tokens, where each transformer encoder layer includes a multi-head self-attention layer and an feed forward network. Inspired by ViT~\cite{dosovitskiy2020image} adding an extra learnable classification token [CLS] to be taken as image representation, we make use of an additional [BODY] token to be served as an abstract representation of the body language and feed it into the transformer encoder along with other tokens. The [BODY] token is randomly initialized before training and jointly optimized with the whole model during training, and its state at the output of the transformer encoder is leveraged to predict the body language vector via a single linear layer. The body language vector $\boldsymbol{l} \in \mathbb{R}^3$ is formulated as follows,
\begin{equation}\label{eq4}
\begin{aligned}
\boldsymbol{E} &= Trans([[BODY], \boldsymbol{M}_{body}^{(1,1)}, \boldsymbol{M}_{body}^{(1,2)}, ..., \boldsymbol{M}_{body}^{(H,W)}]),\\
\boldsymbol{l} &= L2Norm(FC(\boldsymbol{E}^{(1)})),
\end{aligned}
\end{equation}
where $Trans(\cdot)$, $FC(\cdot)$ and $L2Norm(\cdot)$ represents the transformer encoder, linear layer and L2 normalization, respecitvely. 

To facilitate the model to learn the body language vector $\boldsymbol{l}$, we apply a regression loss $loss_{reg}$ to directly optimize the cosine distance between the body language vector and the vector from the sender to the referent, which will be introduced in section~\ref{sec:loss}.

\subsection{Relation Reasoning}\label{sec:RR}
In this section, we perform relation reasoning between the sender and the objects from the sender's perspective by utilizing the spatial coordinates, nonverbal gesture information, and verbal cues. First, to locate the spatial area where the sender faces, we compute the \textbf{spatial attention} between the learned body language vector and spatial coordinates of all the pixels in the image. Then, in that area, we perform the \textbf{noverbal gesture attention} and \textbf{verbal fusion} to find the precise region where the sender points to and describes. Finally, according to the precise region, we generate a box prediction to the referent.

\subsubsection{Spatial Attention}\label{sec:sa} 
The body language vector $\boldsymbol{l}$ defined in section~\ref{sec:BLR} represents the direction from the sender to the referent, revealing an area where the referent might locate. To find and represent the region, we directly compute a spatial attention map $\boldsymbol{A}_{spatial} \in \mathbb{R}^{H_I \times W_I}$ on the image via the cosine similarities between the body language vector $\boldsymbol{l} \in \mathbb{R}^3$ and the spatial coordinates of pixels. The attention score $\boldsymbol{A}_{spatial}(x,y)$ at the position $(x,y)$ in the image is computed as follows,
\begin{equation}
    \boldsymbol{A}_{spatial}(x,y) = \boldsymbol{l} \cdot L2Norm(\boldsymbol{P}_r(x,y)), \\
\end{equation}
where $\boldsymbol{P}_r \in \mathbb{R}^{H_I \times W_I \times 3}$ is the embodied coordinate map defined in the section~\ref{sec:E3CSC}. With the help of the attention map $\boldsymbol{A}_{spatial}$, the noverbal gesture attention and the verbal fusion can be performed in that activated area where referent might locate.

\subsubsection{Nonverbal Gesture Attention}\label{sec:nga} Based on the sender's pointing gesture, the specific region of the referent that the sender points to can be located. To find the specific region, modeling the relations between the sender and regions in the image is not enough, we also need to model the relations among different regions. Without the modeling, the specific region cannot be differentiated from other regions on the same direction that the sender points to. Therefore, we model the relations among the sender and all the regions in the activated area. Similar to the relation modeling in section~\ref{sec:BLR}, we also utilize the transformer encoder to model the relations, which is formulated as follows,
\begin{equation}
\begin{aligned}
    \boldsymbol{M}_{gesture} &= \boldsymbol{M} \odot ReLU(AvgPool(\boldsymbol{A}_{sender} + \boldsymbol{A}_{spatial})), \\
    \boldsymbol{A}_{gesture} &= Softmax(Trans([\boldsymbol{M}_{gesture}^{(1,1)}, \boldsymbol{M}_{gesture}^{(1,2)}, ..., \boldsymbol{M}_{gesture}^{(H,W)}])) 
\end{aligned}
\end{equation}
where $\boldsymbol{A}_{sender}$ and $\boldsymbol{A}_{spatial}$ refer to the sender's region and the activated regions of spatial attention map, respectively, $ReLU(\cdot)$ is the ReLU activation funciton~\cite{agarap2018deep}, $\boldsymbol{M}$ is the multimodal feature map defined in section~\ref{sec:BLR}, and $Softmax(\cdot)$ is the softmax activation function. The gesture attention map  $\boldsymbol{A}_{gesture}\in \mathbb{R}^{H \times W \times 1}$ refers to the specific region of the referent that the sender points to.
Moreover, we propose an attention loss $loss_{attn}$ to facilitate the model to learn the gesture attention map, which is given in section~\ref{sec:loss}.

\subsubsection{Verbal Fusion}\label{sec:vbf}
With the cooperation of gesture, which specifies the specific region of the referent, verbal cues can locate the complete referent. Verbal cues provide straightforward and informative cues and are crucial for reference understanding. Therefore, we utilize the language description to locate the referent in the activated area $\boldsymbol{A}_{spatial}$. Given the language description with $T$ words, we extract the language features $\boldsymbol{L} \in \mathbb{R}^{T \times C}$ from a pretrained BERT~\cite{devlin2018bert} model. Then, we extract the multimodal feature map $\boldsymbol{M}$ because the informative language cues usually describe multimodal information, such as semantic category, visual appearance, and relative spatial location of the referent.
Next, we fuse the language features into the multimodal feature map $\boldsymbol{M}$ to get the verbal-visual feature map  $\boldsymbol{M}_{verbal}\in \mathbb{R}^{H \times W \times C}$. Following ReSC~\cite{yang2020improving}, we use $FiLM$ module~\cite{perez2018film} as
the fusion block, and the feature map $\boldsymbol{M}_{verbal}$ is computed as follows,
\begin{equation}
\begin{aligned}
    \boldsymbol{M}_{verbal} &= FilM(\boldsymbol{M} \odot ReLU(AvgPool(\boldsymbol{A}_{spatial})), Query(\boldsymbol{L})), \\
\end{aligned}
\end{equation}
where the $Query(\cdot)$ is the sub-query learner~\cite{yang2020improving}. We stack three $FiLM$ blocks for verbal fusion following YouRefIt\cite{chen2021yourefit}. Note that the spatial attention map $\boldsymbol{A}_{spatial}$ forces the verbal fusion focus on the activated area. 

Finally, we fuse the nonverbal gesture attention map $\boldsymbol{A}_{gesture}$ to the verbal-visual feature map $\boldsymbol{M}_{verbal}$ to predict the anchor boxes and their corresponding confidence scores. The fusion is implemented via a concatenation operation followed by a stack of convolutional layers.

\subsection{Loss Function}\label{sec:loss}
\subsubsection{Regression Loss} We calculate $\boldsymbol{p}_{box} \in \mathbb{R}^3$ by averaging emobodied 3D coordinates of pixels in the bounding box of ground-truth referent and take it as supervision to learn the body language vector $\boldsymbol{l} \in \mathbb{R}^{3}$. The regression loss $loss_{reg}$ is computed as follows:
\begin{equation}\label{eq8}
loss_{reg} = 1-L2Norm({\boldsymbol{p}_{box}}) \cdot \boldsymbol{l}.
\end{equation}
\subsubsection{Attention Loss} The attention loss $loss_{attn}$ is computed between the learned nonverbal gesture attention map $\boldsymbol{A}_{gesture}$ and the ground-truth bounding box $box \in \mathbb{R}^{H \times W}$ as follows,
\begin{equation}\label{eq9}
loss_{attn} = 1- \sum_{x,y=1}^{H, W} \boldsymbol{A}_{gesture}(x,y) \ast box(x,y),
\end{equation}
where $box(x,y)=1$ if the position $(x,y)$ is in the ground-truth bounding box; otherwise $box(x,y)=0$.

\subsubsection{Overall Loss} Following YouRefIt~\cite{chen2021yourefit}, we apply the diverse loss~\cite{yang2020improving} and the YOLO's loss~\cite{redmon2018yolov3} to jointly optimize the model. 
Finally, our loss function can be calculated as follows,
\begin{equation}\label{eq10}
loss = loss_{yolo}+loss_{div}+loss_{reg} + loss_{attn}.
\end{equation}
The diverse loss enforces the diversity of words in different rounds. It is formulated as $loss_{d i v}=\left\|A^{T} A \odot(\mathbf{1}-I)\right\|_{F}^{2}$, where $A$ is the attention score matrix in the sub-query module~\cite{yang2020improving} and $I$ is an identity matrix.

\section{Experiments}

\subsection{Dataset and Evaluation Metric}
We evaluate the proposed REP on the released indoor image benchmark YouRefIt~\cite{chen2021yourefit} for Embodied Reference Understanding task (ERU). Note that the video version of YouRefIt is not released when this paper submits. YouRefIt$\footnote[1]{\url{https://github.com/yixchen/YouRefIt\_ERU}}$ contains $4221$ query-referent pairs with $395$ object categories. It is split into train and test, which has $2970$ and $1251$ samples, respectively. The average length of language descriptions is $3.73$ and extra nonverbal cues such as gesture and orientation are provided. 
The Prec@X metric is used to evaluate the performance of ERU models on different sizes of referents and the overall performance. The $Prec@X$ is the percentage of prediction bounding boxes whose IoU scores are higher than a given threshold $X$, where $X\in \{0.25,0.50,0.75\}$.

\subsection{Implementation Details}
\noindent\textbf{Networks Architecture.} For a fair comparison with
previous works~\cite{yang2020improving,chen2021yourefit}, we adopt Darknet-53~\cite{redmon2018yolov3} pretrained on MSCOCO object detection dataset~\cite{lin2014microsoft} as the visual backbone. Language features are encoded by BERT-base~\cite{devlin2018bert} followed by two fully connected layers.
Following ReSC-large~\cite{yang2020improving}, we keep the ratio of height and width and resize the long edge of the input image to $512$. Then, we pad the resized image to $512\times 512$, \ie, $H_I=W_I=512$. And the $H$, $W$, and $C$ are $32$, $32$ and $256$, respectively. The number of transfomer encoder layers are $2$. For each batch, we randomly sample $16$ sentences and images with random horizontal flip, random intensity, saturation change, and random affine transformation following previous works~\cite{yang2019fast,yang2020improving}. We Adopt the RMSProp~\cite{tieleman2012lecture} optimizer with weight decay $0.0005$. The initial learning rate is set to $0.0001$ and reduced by half every $10$ epochs for a total of $100$ epochs. The weights of each loss are set to be 1. All the experiments are implemented in PyTorch~\cite{paszke2019pytorch}, with the NVIDIA GeForce RTX 3090. 

\subsection{Comparison with State-of-the-Arts}
\begin{table*}[htb]
\caption{Comparison with state-of-the-art methods on YouRefIt dataset. The best performing method is marked in bold.}
\centering
\label{table1}
\resizebox{\textwidth}{!}{
\begin{tabular}{@{}lcccccccccccc@{}}
\toprule
\multicolumn{1}{c}{}                        & \multicolumn{4}{c}{IoU=0.25}                                  & \multicolumn{4}{c}{IoU=0.50}                                  & \multicolumn{4}{c}{IoU=0.75}                                  \\ \cmidrule(l){2-13} 
\multicolumn{1}{c}{\multirow{-2}{*}{Model}} & all           & small         & medium        & large         & all           & small         & medium        & large         & all           & small         & medium        & large         \\ \midrule                                                                       
$\rm{MAttNet_{pretrain}}$~\cite{yu2018mattnet}                           & 14.2          & 2.3           & 4.1           & 34.7          & 12.2          & 2.4           & 3.8           & 29.2          & 9.1           & 1.0           & 2.2           & 23.1          \\
$\rm{FAOA_{pretrain}}$                                   & 15.9          & 2.1           & 9.5           & 34.4          & 11.7          & 1.0           & 5.4           & 27.3          & 5.1           & 0.0           & 0.0           & 14.1          \\
$\rm{ReSC_{pretrain}}$                              & 20.8          & 3.5           & 17.5          & 40.0          & 16.3          & 0.5           & 14.8          & 36.7          & 7.6           & 0.0           & 4.3           & 17.5          \\                                                                      
FAOA~\cite{yang2019fast}                                             & 44.5          & 30.6          & 48.6          & 54.1          & 30.4          & 15.8          & 36.2          & 39.3          & 8.5           & 1.4           & 9.6           & 14.4          \\
ReSC~\cite{yang2020improving}                                             & 49.2          & 32.3          & 54.7          & 60.1          & 34.9          & 14.1          & 42.5          & 47.7          & 10.5          & 0.2           & 10.6          & 20.1          \\
$\rm{YouRefIt_{PAF\_only}}$                          & 52.6          & 35.9          & 60.5          & 61.4          & 37.6          & 14.6          & 49.1          & 49.1          & 12.7          & 1.0           & 16.5          & 20.5          \\
$\rm{YouRefIt_{Full}}$~\cite{chen2021yourefit}                                   & 54.7          & 38.5          & 64.1          & 61.6          & 40.5          & 16.3          & 54.4          & 51.1          & 14.0          & 1.2           & 17.2          & 23.2          \\
Ours $\rm{REP_{Full}}$                                 & \textbf{58.8} & \textbf{44.7} & \textbf{68.9} & \textbf{63.2} & \textbf{45.7} & \textbf{25.4} & \textbf{57.7} & \textbf{54.3} & \textbf{18.8} & \textbf{3.8}           & \textbf{22.2} & \textbf{29.9} \\ \bottomrule
\end{tabular}
}
\end{table*}
We compare our model with baselines and state-of-the-art methods on ERU, including MattNet~\cite{yu2018mattnet}, FAOA~\cite{yang2019fast}, ReSC~\cite{yang2020improving} and YouRefIt~\cite{chen2021yourefit}. Experimental results are shown in Table~\ref{table1}. Our REP consistently outperforms all the state-of-the-art models (SOTAs) across all the indicators by large margins. REP improves the average performance of $Prec@0.25$, $Prec@0.50$ and $Prec@0.75$ achieved by the existing best method by $4.1\%$, $5.2\%$ and $4.8\%$, respectively.

Compared with the models pretrained on traditional REC dataset~\cite{yu2016modeling}, our REP achieves $29.4\%$ improvements in terms of $Prec@0.50$ and $26.2\%$ in average, which demonstrates the significant difference between REC task and ERU task. Compared with FAOA and ReSC, REP improves the $Prec@0.25$, $Prec@0.50$, and $Prec@0.75$ by $9.6\%$, $10.8\%$ and $8.3\%$, respectively, which reveals the importance of nonverbal cues in ERU. 

Our REP significantly surpasses YouRefIt by $4.7\%$ on average of all indicators, although YouRefIt already inputs nonverbal cues (\ie, part affinity field map and saliency map) for multimodal fusion. The comparison demonstrates the effectiveness of our 3D view rotation and relation reasoning for addressing the perspective-taking challenge in ERU. Note that REP improves more on the more challenging referring of small objects. Thanks to the relation reasoning of REP, it improves small objects' grounding accuracy $Prec@0.25$ and $Prec@0.50$ by $6.2\%$ and $9.1\%$, respectively.

\subsection{Ablation Study}
We conduct an ablation study to evaluate the effectiveness of 3D view rotation and relation reasoning methods, and the results are shown in Table~\ref{table2}.
\begin{table}[ht]
\caption{Ablation study of 3D view rotation and relation reasoning methods. The best performing method is marked in bold.}
\label{table2}
\centering
\resizebox{1.0\textwidth}{!}{
\begin{tabular}{@{}llcccccccccccc@{}}
\toprule
\multicolumn{2}{c}{} & \multicolumn{4}{c}{IoU=0.25}   & \multicolumn{4}{c}{IoU=0.50}   & \multicolumn{4}{c}{IoU=0.75}   \\ \cmidrule(l){3-13} 
\multicolumn{2}{c}{\multirow{-2}{*}{Model}}                       & all  & small & medium & large & all  & small & medium & large & all  & small & medium & large \\ \midrule
1& baseline               & 54.3   & 39.7   & 60.7  & 62.6 & 39.0  & 18.4  & 48.6    & 50.0 & 11.0    & 2.3  & 9.1   & 20.7  \\ \midrule
2&+depth estimation      & 56.4 & 42.1  & 62.0   & 65.1  & 40.8 & 19.6  & 50.4   & 52.5  & 12.3 & 2.8   & 11.1   & 22.4  \\
3&+embodied coordinate   & 56.7 & 42.3  & 62.5   & 65.1  & 41.7 & 20.3  & 51.7   & 53.2  & 14.5 & 3.4   & 14.6   & 24.9  \\
4&+body language vector  & 57.1 & 44.4  & 64.0   & 63.0  & 42.4 & 23.2  & 52.6   & 51.6  & 16.1 & 3.1   & 18.7   & 26.0  \\ \midrule
5&+verbal attention  & 57.7 & 44.0  & 63.3   & \textbf{65.9}  & 44.0 & 23.2  & 54.2   & \textbf{54.6}  & 18.0 & 3.8   & 20.0   & 30.0   \\
6&+gesture attention         & \textbf{58.3} & \textbf{44.7} & \textbf{68.9} & 63.2 & \textbf{45.7} & \textbf{25.4} & \textbf{57.7} & 54.3 & \textbf{18.8} & \textbf{3.8}           & \textbf{22.2} & \textbf{30.0}  \\ \bottomrule
\end{tabular}
}
\end{table}

\textbf{Baseline}. Baseline model shares the same visual encoder Darknet-53~\cite{redmon2018yolov3} and textual encoder BERT~\cite{devlin2018bert} with our REP and also cooperates with nonverbal and verbal cues for prediction. It first obtains the multimodal feature map by fusing the visual feature map $\boldsymbol{S}_v$, 2D image coordinates, and part affinity field map $\boldsymbol{S}_{gesture}$, and then fuses verbal representation $Query(\boldsymbol{L})$ to the multimodal feature map via a stack of three FiLM layers~\cite{perez2018film,yang2020improving}. REP improves $Prec@0.25$, $Prec@0.50$, and $Prec@0.75$ of baseline by $4.0\%$, $6.7\%$ and $7.8\%$, respectively.

\textbf{3D View Rotation}. As shown in line 2, +depth estimation improves the grounding accuracy by $1.9\%$, $1.8\%$ and $1.3\%$ in terms of $Prec@0.25$, $Prec@0.5$, and $Prec@0.75$, respectively, which demonstrates that the depth information can help to locate the referent. The cooperation of depth estimation and embodied coordinate (line 3) improves baseline by $2.4\%$, $2.7\%$ and $3.5\%$ in terms of $Prec@0.25$, $Prec@0.50$, $Prec@0.75$, respectively, which shows the effectiveness of rotating the receiver to the position of the sender to construct the embodied 3D coordinate system. The body language vector (line 4 vs. line 3) further significantly improves average accuracy on $Prec@0.75$ by $1.6\%$. The reason is that the body language vector encodes the body orientation and gesture of the sender, which could indicate the rough area where the referent locates. In general, the 3D view rotation module outperforms the baseline by $2.8\%$, $3.4\%$, and $5.1\%$ in terms of $Prec@0.25$, $Prec@0.50$, and $Prec@0.75$, respectively. 

\textbf{Relation Reasoning}. The verbal attention aims to cooperate the spatial attention to locate the referent in the activated area where the sender faces. With verbal attention, the model (line 5 vs. line 4) improves the overall grounding accuracy $Prec@0.25$ and $Prec@0.5$ by $1.6\%$ and $1.7\%$, respectively. The improvement shows that the verbal attention method helps utilize verbal cues better. The nonverbal gesture attention aims to find the specific region where the sender points to and uses the specific region to help locate the complete referent by cooperating with verbal attention. The model with nonverbal gesture attention (line 6 vs. line 5) achieves $1.7\%$ significant improvement in terms of $Prec@0.50$, and it improves more for locating referents with small and medium sizes. In detail, with nonverbal gesture attention, the $Prec@0.5$ of the model for finding small and medium referents is improved by $2.2\%$ and $3.5\%$, respectively. 

\subsection{Qualitative Evaluation}
\begin{figure*}[t]
\centering
\subfloat[]{
\includegraphics[width=0.18\textwidth]{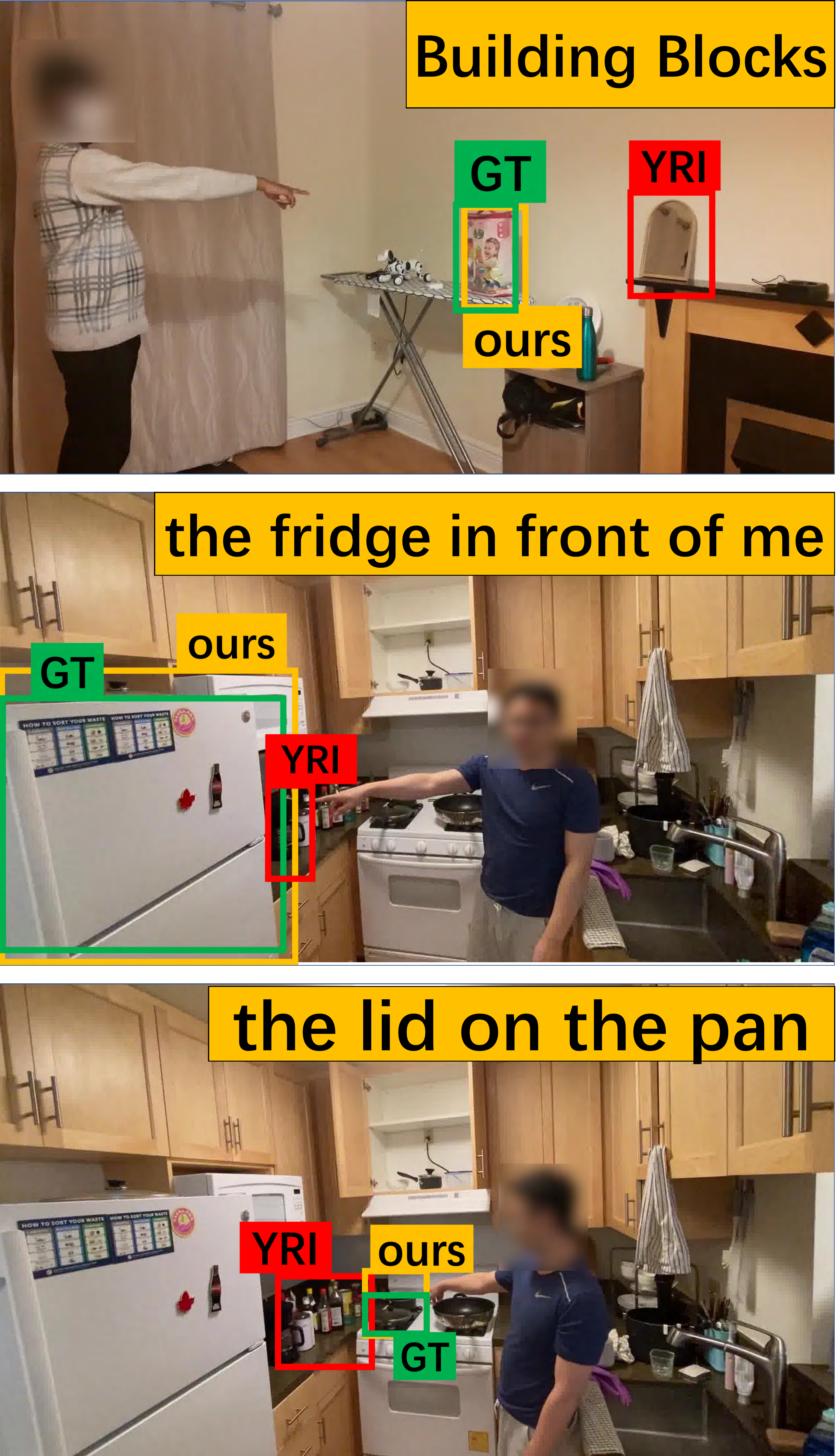}
}
\subfloat[]{
\includegraphics[width=0.18\textwidth]{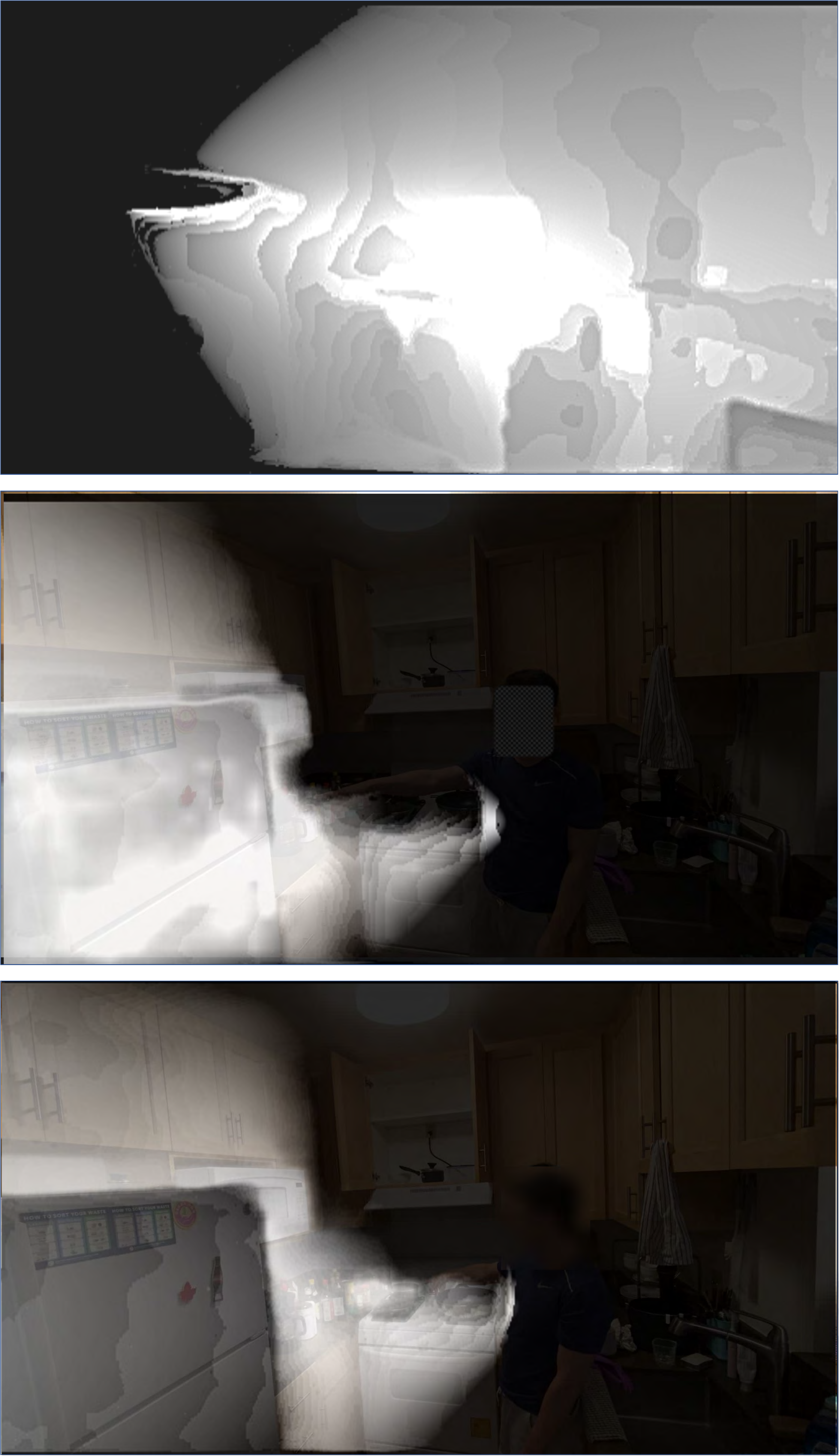}
}
\subfloat[]{
\includegraphics[width=0.18\textwidth]{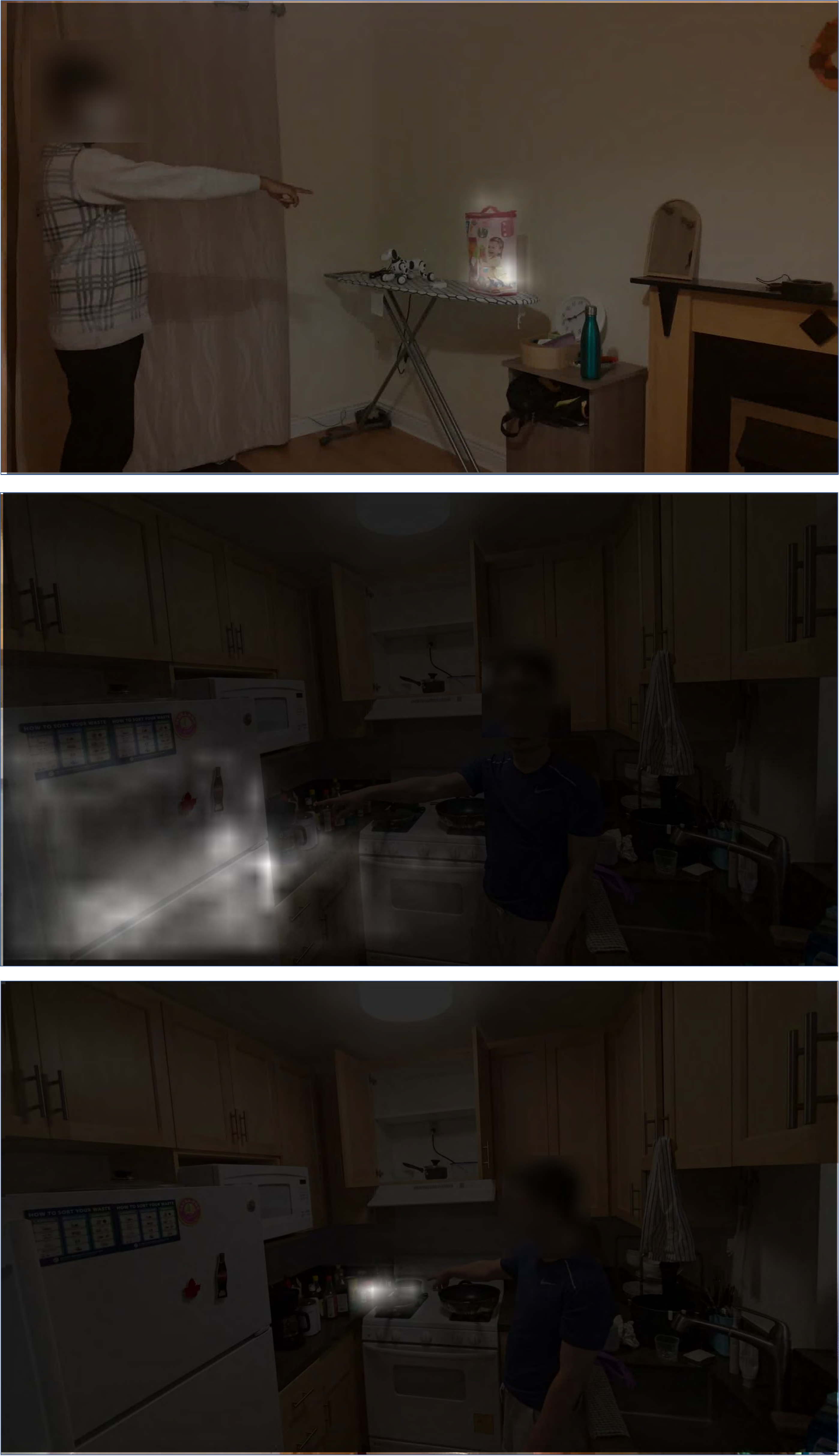}
}
\subfloat[w/]{
\includegraphics[width=0.18\textwidth]{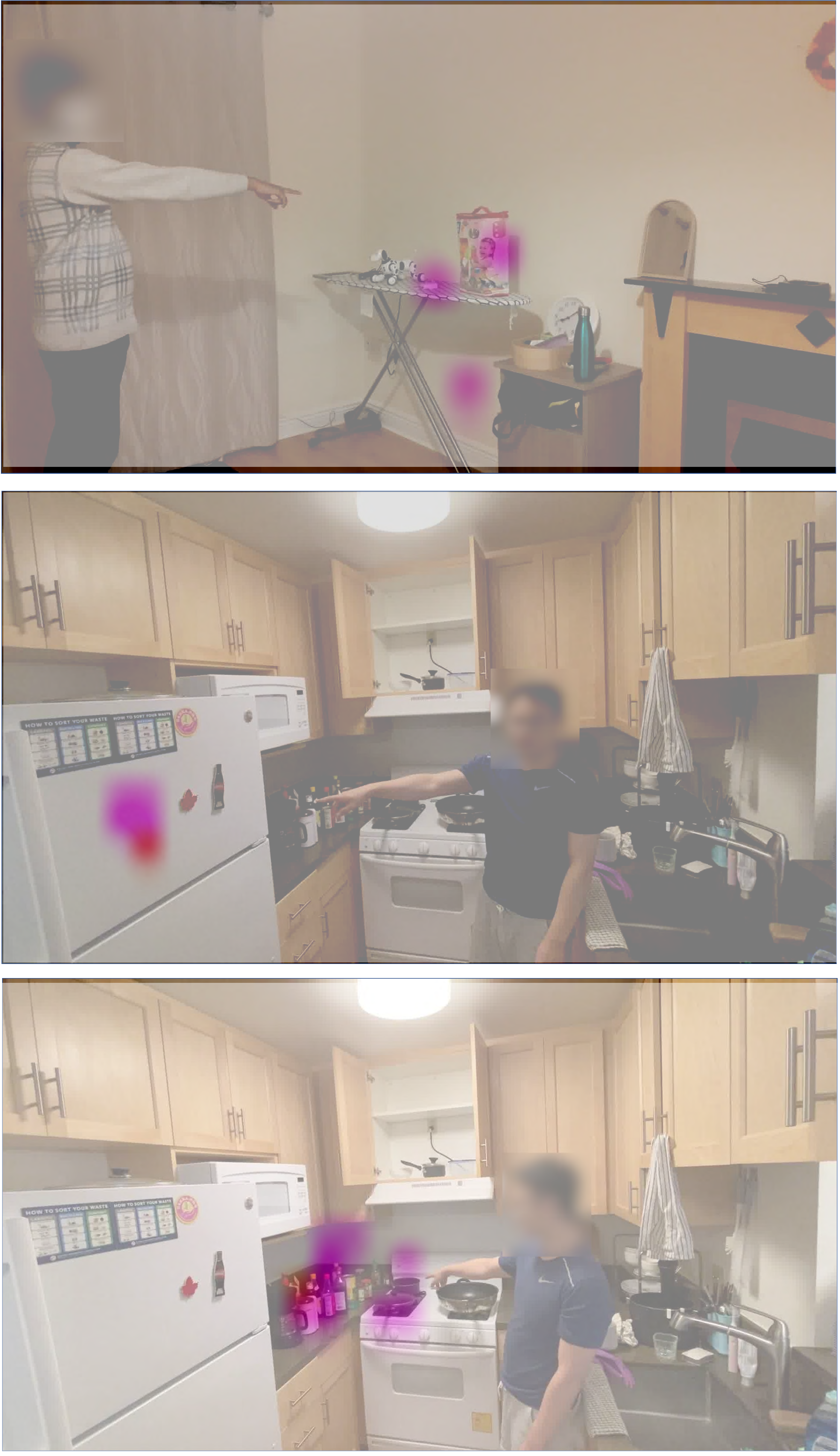}
}
\subfloat[w/o]{
\includegraphics[width=0.18\textwidth]{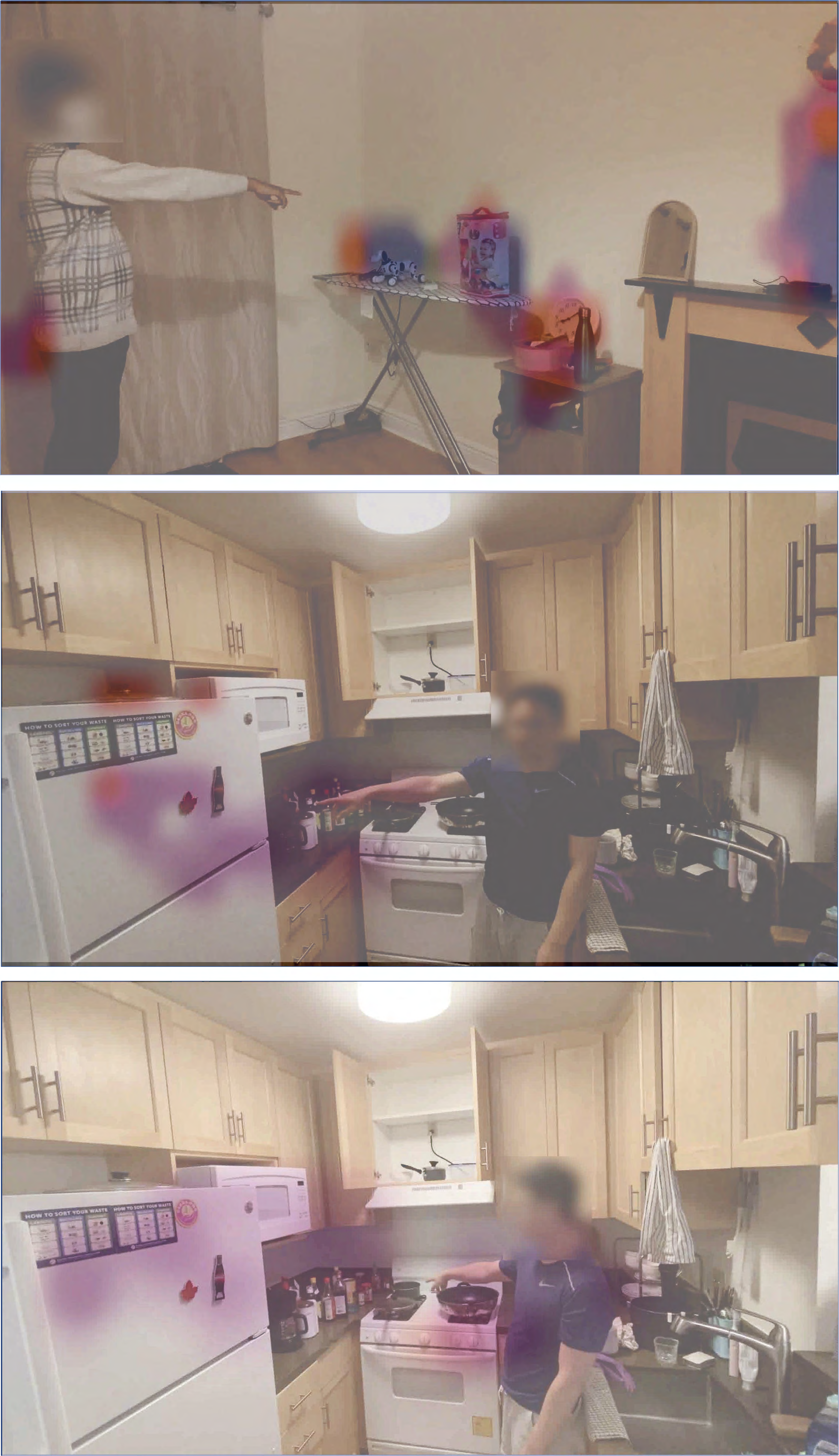}
}
\caption{Qualitative Results showing (1) prediction results: green, yellow and red boxes are the ground-truth, our prediction result, and YouRefIt's predicted referents; (2) spatial attention; (3) gesture attention; (4) verbal fusion heatmap with the help of attention map; (5) verbal fusion headmap without the help of attention map.}
\label{fig:vis}
\end{figure*}

To better explore in-depth insights into the view rotation and relation reasoning based on the embodied 3D coordinate system, we visualize three examples along with prediction results, spatial attention, nonverbal attention, and verbal attention maps. The visualization is shown in Fig.\ref{fig:vis}. Two different verbal attention maps are visualized to show the effect of with or without the help of the spatial attention map for the verbal fusion. Following~\cite{yang2020improving}, we use confidence scores to represent the verbal attention scores by adopting an output head over the verbal-visual feature map at the last layer.

As shown in Fig.\ref{fig:vis}, REP can generate the explainable visual evidence of stepwise reasoning from the spatial attention to the nonverbal gesture attention and the verbal attention, and locates the referent from the sender's perspective in different kinds of challenging scenarios. (1) In the first example, REP successfully finds the activated area where the sender faces and locates the ``building blocks'' while excluding the distractor ``bag''. (2) Thanks to the view rotation, our REP precisely captures the slight differences in the sender's body orientation and generates distinct activated areas for the second and third examples. (3) With the help of spatial attention, the verbal fusion module locates the referent accurately for the novel object of ``the lid on the pan". (4) The nonverbal gesture attention module and verbal fusion module can cooperate in locating the referent. The nonverbal gesture attention module finds a specific region of ``the fridge in front of me", and the verbal fusion module helps to locate the referent completely.

\section{Conclusion}
In this paper, we propose Reasoning from your Respective (REP) model to tackle the Embodied Reference Understanding (ERU) task. REP first rotates the receiver to the position of the sender and estimate the sender's viewpoint to the referent by constructing the embodied 3D coordinate system and learning the body language representation. Then, REP performs relation reasoning between the sender and the referent by cooperating the spatial attention, nonverbal gesture attention and the verbal fusion methods. REP not only outperforms the state-of-the-art models of ERU by a large margin but also generates explainable visual evidence of step-by-step reasoning.
\\

\textbf{Acknowledgment} This work is supported by Shanghai Pujiang Program (No.21PJ1410900).
\clearpage
%
%
\bibliographystyle{splncs04}
\bibliography{egbib}
\end{document}